\documentclass{comnet}

\usepackage{amssymb}
\usepackage{amsmath}
\usepackage{algorithm}
\usepackage{algpseudocode}
\usepackage{subcaption}
\usepackage{hyperref}
\usepackage{afterpage}
\usepackage{graphicx}
\graphicspath{{figures/}}

\begin{document}

\title{MSGCN: Multiplex Spatial Graph Convolution Network for Interlayer Link Weight Prediction}

\shorttitle{MSGCN for Interlayer Link Weight Prediction} 
\shortauthorlist{S. Wilson, S. Khanmohammadi} 

\author{
\name{Steven E. Wilson$^{a}$}
\address{$^{a}$Data Science and Analytics Institute, University of Oklahoma, 202 W. Boyd St., Norman, Oklahoma, 73019, United States}
\name{Sina Khanmohammadi$^{a,b,*}$}
\address{$^{a}$Data Science and Analytics Institute, University of Oklahoma, 202 W. Boyd St., Norman, Oklahoma, 73019, United States}
\address{$^{b}$School of Computer Science, University of Oklahoma, 110 W. Boyd St., Norman, Oklahoma, 73019, United States
\email{$^*$Corresponding author: sinakhan@ou.edu}}
}

\maketitle


\begin{abstract}
{Graph Neural Networks (GNNs) have been widely used for various learning tasks, ranging from node classification to link prediction. They have demonstrated excellent performance in multiple domains involving graph-structured data. However, an important but less explored learning task is link weight prediction which is more complex than binary link classification. Link weight prediction becomes even more challenging when considering multilayer networks, where nodes can be connected across multiple layers. 
To address these challenges, we propose a new method called Multiplex Spatial Graph Convolution Network (MSGCN), which spatially embeds information across multiple layers to predict interlayer link weights. The MSGCN method generalizes spatial graph convolution to multiplex networks and captures the geometric structure of nodes across
multiple layers.  Extensive experiments using data with known interlayer link information show that the MSGCN model has robust, accurate, and generalizable link weight prediction performance across a wide variety of network structures. We also demonstrate a real‑world application of the proposed method using the London transportation network. In this setting, MSGCN accurately predicts passenger loads in the multiplex network, where the interlayer link weights represent the number of passengers traveling between stations that are not directly connected.}
{Multiplex Spatial Networks, Link Weight Prediction, Graph Convolution}
\end{abstract}

\section{Introduction}
\label{sec1}

Many engineered and natural systems are composed of interconnected nodes in a multilayer network \cite{Kivela_2014,boccaletti2014structure,aleta2019multilayer,de2023more}. Multilayer networks can arise in various systems, such as transportation \cite{aleta2017multilayer,alessandretti2023multimodal}, communication \cite{lehman2011multilayer,wu2020traffic}, social \cite{finn2021multilayer,dragic2021multilayer}, and biological systems \cite{vaiana2020multilayer,liu2020robustness}. One of the key elements that could help us understand the behavior of such complex multilayer networks is the interlayer relationship between the nodes. In this regard, several methods have been developed for multilayer link prediction, where the goal is to identify the presence or absence of a link between nodes at different layers of a network \cite{yang2024link}. Examples of such methods include using topological properties of networks, including motifs and triadic closures \cite{aleta2020link, gao2023novel, liu2024motifs}, utilizing stochastic processes, particularly random walks on graphs \cite{Liu_2017, baptista2022universal, liu2024tlfsl, luo2024link}, employing information-theoretic measures \cite{jafari2021information}, decision fusion-based models such as mixture of experts \cite{la2025heuristic, wangmultiple2025}, and, more recently, deep learning-based approaches based on graph neural networks \cite{ren2024link, zangari2024link}.

Despite the encouraging results from these methods, several challenges remain. First, while link prediction models provide important information about the existence of interlayer links, they do not offer insights into the strength of these connections, which is essential for understanding overall system behavior \cite{lu2011link,ding2024wepred}. Second, most of these methods do not scale well to large graphs, particularly graph neural network-based approaches, which suffer from oversmoothing \cite{rusch_2023,wu2023demystifying}. Lastly, the multilayer networks in these methods do not preserve node order. Consequently, they do not consider the spatial information embedded in the nodes, which is crucial in real-world applications such as transportation networks, where the location of cities matter \cite{iddianozie2020improved, zhu2022spatial}.

In this paper, we propose a new method for interlayer link weight prediction in multiplex networks that addresses these challenges. The proposed method, called Multiplex Spatial Graph Convolution Network (MSGCN), uses node projection across layers combined with graph convolution to learn interlayer node dependencies in a multiplex network. The method also maintains the order of nodes by taking into account the spatial information of the nodes during the convolution process. Lastly, our proposed method addresses the oversmoothing problem by using a custom loss function for the learning process. We demonstrate the effectiveness of our method using synthetic data with known interlayer links and showcase its robustness and generalizability to various multiplex networks with different sizes and topologies. We further show a practical application of the proposed method using the London transportation network data, which includes the multiplex representations of traveler movements on the London Underground from 2016 to 2023.

The rest of the paper is organized as follows: In Section \ref{Preliminaries}, we provide the preliminary information and introduce the notations used in this study. In Section \ref{Methodology}, we explain the details of our proposed MSGCN method for spatial graph convolution. Section \ref{Results} presents the results of our method, followed by a discussion and conclusion in the subsequent sections.

\section{Preliminaries}
\label{Preliminaries}

In this section, we provide the preliminary information needed to introduce our proposed method. We first introduce a mathematical representation of multilayer networks followed by a description of single layer spatial graph convolution that is the basis of our model.  

\subsection{Multilayer Networks}

We start with a single layer network $G$, which can be formalized as a weighted graph, $G = (V, E, w)$ where $V$ is the set of vertices (nodes), $E = V \times V$ is the set of edges (links) connecting vertices, and $w: E\rightarrow \mathbb{R}$ is the weight function that assigns a weight to each edge. The connectivity matrix, ${\bf C}$, represents the network structure where each element of the matrix is ${\bf C}_{ij} = w(v_i, v_j)$. The multilayer network generalizes the single layer network by adding a labeling function $\ell:V \rightarrow \mathbb{Z}$, which maps each vertex $v$ to a layer index $l$. Each layer represents a different view (i.e., aspect or dimension) of the system. The multilayer network representation then becomes $M = (V, E, w, \ell)$, where each edge from node $v_i$ in layer $\ell(i)=p$ to node $v_j$ in layer $\ell(j)=q$ is represented by $e_{ip,jq}=(v_{i}, p,v_{j}, q, w(v_{ip}, v_{jq}))$. 

\begin{figure}[!ht]
     \centering
     \includegraphics[width=\textwidth,height=0.25\textheight,keepaspectratio]{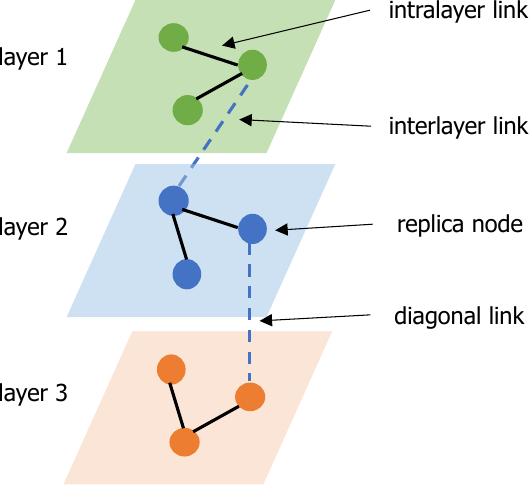}
     \caption{Sample spatial multiplex network showing different link and node types between layers.}
     \label{fig:sample-network}
 \end{figure}

The links within the same layer $\ell(i) = \ell(j)$ are called intralayer links and represent the relationship between nodes within the same layer. Conversely, links spanning across layers $\ell(i) \neq \ell(j)$ are called interlayer links and capture the interactions between nodes in different layers. The interlayer links connecting the same nodes across different layers (replica nodes) are called diagonal links. It should be noted that if we only have the same set of vertices in each layer, the multilayer network is called a multiplex network. If each vertex includes spatial information (such as coordinates), it is called a spatial multiplex network. An example of a simple spatial multiplex network is shown in Figure \ref{fig:sample-network}.

\subsection{Spatial Graph Convolution}
Graph Convolutional Networks (GCNs) have been shown to be effective in deep learning tasks on graph structured data for node classification \cite{kipf_2017, hamilton_2018}, link prediction \cite{kipf_2016}, and graph classification \cite{Wu_2019}. However, conventional GCN models are unable to preserve the spatial relationships of nodes since positional information is lost during the convolution operation \cite{danel_2020}. To address this limitation, Spatial Graph Convolutional Networks (SGCNs) have been proposed to preserve node positions during graph convolution \cite{danel_2020}. In spatial graph convolutional networks, positions for each node, ${\bf p}_i$, are preserved across convolutional layers. To accomplish this, the standard message passing in graph convolutional networks is modified as follows:

\begin{equation}
    \overline{\textbf{h}}_i(\textbf{U}, \textbf{b}) = \sum_{j \in N_i} \phi (\textbf{U}^T(\textbf{p}_j - \textbf{p}_i) + \textbf{b}) \odot \textbf{h}_j  
    \label{eq:sgcn}    
\end{equation}
where $\mathbf{U}$ and $\mathbf{b}$ are the trainable parameters and bias vectors, respectively. $N_i$ is the connected neighbors to node $i$, $\mathbf{p}$ is the node position vector, and $\mathbf{h}$ is the feature vector. Finally, $\phi$ is a non-linear transform function and $\odot$ is the element-wise multiplication resulting in the embedded feature vector ($\overline{\mathbf{h}}_i$) \cite{danel_2020}.






\section{Methodology}\label{Methodology}

\subsection{The MSGCN Framework}

\begin{figure}[!ht]
\centering
\includegraphics[width=\textwidth]{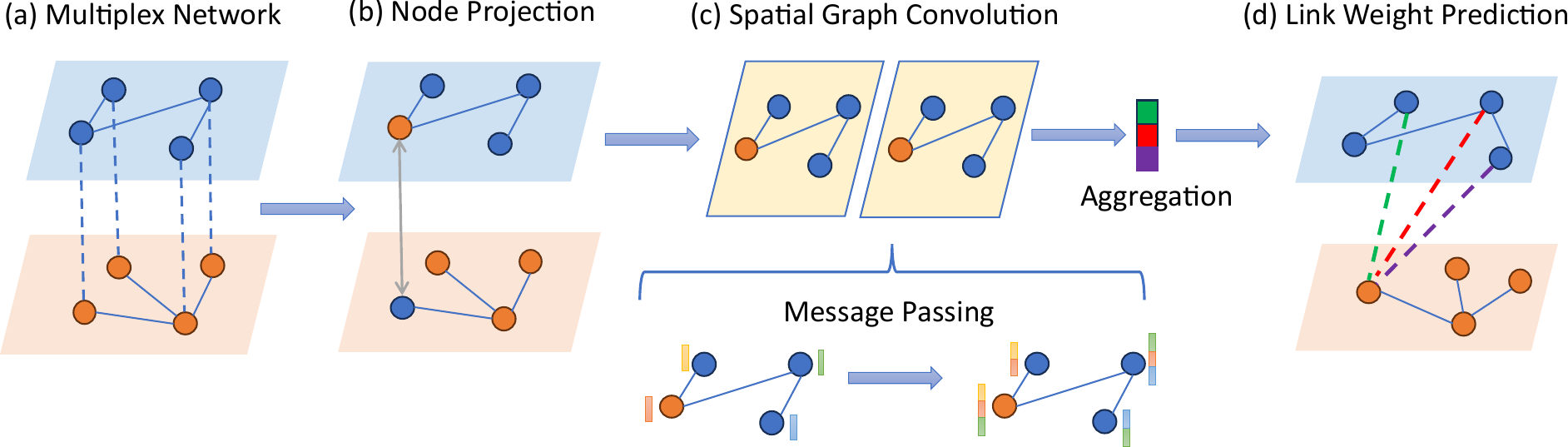}
\caption{Overview of the Multiplex Spatial Graph Convolution Network (MSGCN): The process begins in (a) with a multiplex network where nodes are in the same spatial position across layers. In (b), node features are projected across diagonal links to adjacent layers. Next, in (c), spatial graph convolution is applied to each projected graph.  The spatial graph convolution network layers are shown in yellow to distinguish them from the multiplex network layers.  Message passing captures the projected node's relationship to it's neighbors. Finally, in (d), the aggregation step from the spatial graph convolution estimates the predicted interlay link weights.}
\label{fig:method}
\end{figure}

The overall framework of the Multiplex Spatial Graph Convolution Network (MSGCN) is shown in Figure \ref{fig:method}. We start with a multiplex network and project the features of each node at the same spatial location across different layers. This results in single-layer graphs for each projected node, representing the interlayer links between a node from the original layer and nodes in the layer to which the node was projected. Next we apply spatial graph convolution to each projected graph using the multilayer spatial message passing given by the following equation:

\begin{equation}
    \overline{\textbf{h}}_{(i,l)}(\textbf{U}, \textbf{b}) = \sum_{j \in N_{i}} \phi (\textbf{U}^T(\textbf{p}_{(j,l)} - \textbf{p}_{(i,l)}) + \textbf{b}) \odot \textbf{h}_{(j,l)}  
    \label{eq:multisgcn}    
\end{equation}

This multilayer spatial message passing is an extension of Equation \ref{eq:sgcn}, where $l$ represents a layer in the multilayer network. In our framework, global mean pooling is applied to aggregate the messages from neighbors, and then a ReLU transformation function ($\phi$) is used to produce link weights. These link weights then become the predicted interlayer link weights in our multiplex network.

\subsection{Preventing Oversmoothing}

An inherent limitation of graph convolutional networks is oversmoothing, where node features become more similar as the network size increases \cite{rusch_2023}. To mitigate this effect in our method, we developed a custom loss function designed to maintain the variance and range of predictions. This ensures effective scaling of predictors as the network size increases. A detailed description of the custom loss function is provided here. Given a vector of predicted values $\mathbf{\hat{y}}$ and a vector of target values $\mathbf{y}$, the total loss $L$ is computed as the weighted sum of three individual loss components: mean squared error loss ($L_{\text{MSE}}$), spread loss ($L_{\text{spread}}$), and range penalty ($L_{\text{range}}$). The goal of the proposed loss function (provided in Equation \ref{eq:loss}) is to find the parameters that minimize the prediction error while maintaining the variance and range of the target values.

\begin{equation}
L = w_{\text{MSE}} \cdot L_{\text{MSE}} + w_{\text{spread}} \cdot L_{\text{spread}} + w_{\text{range}} \cdot L_{\text{range}} 
\label{eq:loss}
\end{equation}    

Here, the mean squared error loss is defined as: 
\begin{equation}
L_{\text{MSE}} = \frac{1}{N} \sum_{i=1}^{N} (\hat{y}_i - y_i)^2 \end{equation}
where $\hat{y}_i $ and $y_i$ are the predicted and actual values for the $i$'th sample, and $N$ is the total number of samples.  This first loss term is included to minimize the error between the actual and predicted values. The next component of the loss function (spread loss) is defined as:

\begin{equation}
L_{\text{spread}} = \frac{1}{{\sigma_{\hat{y}}^2} + \epsilon}
\end{equation}
where ${\sigma_{\hat{y}}^2}$ is the variance of the predicted values and $\epsilon$ is a small constant to avoid division by zero.  The second loss term minimizes the precision (inverse of of the variance) in the predicted values, so that they don't converge to the mean value when we have large number of samples. Finally, the last component of the loss function is defined as:
\begin{equation}
L_{\text{range}} = \frac{1}{N} \sum_{i=1}^{N} (\hat{y}_i - {y}_{\text{range}})^2 
\end{equation}
Here, ${y}_{\text{range}}$ is the range of the values in our training set. Hence, the range loss penalizes predictions that are far from the range of the actual values. This ensures that the spread loss introduced previously does not cause the predictions to go beyond the range of actual values.

\subsection{The Experimental Design} \label{sec4}

\subsubsection{Data Description}

In order to evaluate our proposed framework, we used both synthetic data with known inter‑layer link weights and real‑world passenger‑flow data from the London transportation network \cite{london_datastore}. The details of each dataset are described below.

\begin{figure}[!ht]
    \centering
    \includegraphics[width=\textwidth] {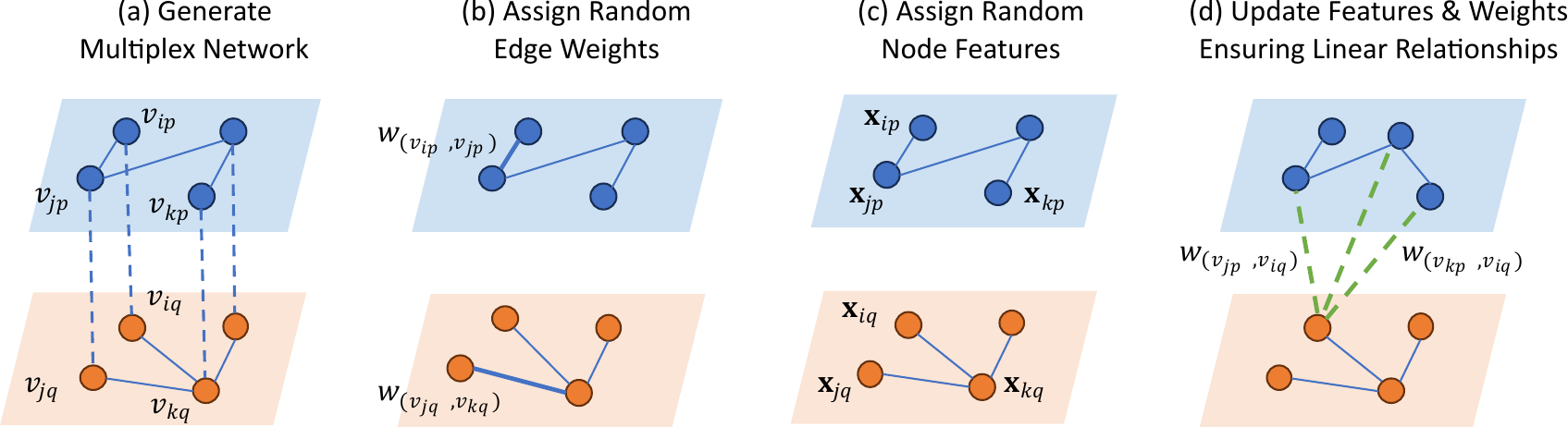}
    \caption{Details of the data generation with only a few representative variables labeled in the figure for clarity.  In (a) an initial spatial multiplex network is created using the parameters in Table \ref{tab:graph_parameters}. Next, in (b) random weights are assigned to each intralayer link. Then, in (c) random node feature values are assigned to each node. Finally, in (d) node features and interlayer link weights are updated using equations \ref{eq:graph_generation} and \ref{eq:edge_weights} to ensure a linear relationships between neighbors and layers.}
    \label{fig:graph-creation}
\end{figure}

\paragraph{Synthetic Data:}
To confirm our methods ability to predict interlayer link weights, we first generated synthetic data with known interlayer link weights using the procedure outlined in Figure \ref{fig:graph-creation}. Step one of the process involves generating a baseline spatial network with the number of nodes and layers outlined in Table \ref{tab:graph_parameters}, where each replica node across all layers is assigned the same random spatial position in the (x, y) coordinates. Next, in step two, a random link weight ${w \in [0,1]}$ is assigned to each intralayer connection between nodes within the same layer. In step three, each node is assigned a random feature value ${{\bf x} \in [0,1]}$. Lastly, in step four, each node is connected to every other node across adjacent layers. The node feature values are then updated according to equation \ref{eq:graph_generation} to ensure a linear relationship between neighboring nodes in the same layer.

\begin{equation}
{\bf x}_{ip} = \sum_{j=1}^{M} w(v_{ip}, v_{jp}) {\bf x}_{jp} \quad \label{eq:graph_generation}  
\end{equation}

Here, ${\bf x}_{ip}$ and ${\bf x}_{jp}$ are the feature values of nodes $i$ and $j$ in layer $p$, and $w(v_{ip}, v_{jp})$ is the weight of the intralayer link from node $i$ to node $j$. The total number of nodes in one layer (i.e., layer $p$) is represented by $M$. Lastly, the interlayer link weights are updated using equation \ref{eq:edge_weights} to ensure a linear relationship exists between nodes in adjacent layers.

\begin{equation}
w(v_{ip}, v_{jq}) = ({\bf x}_{ip} + {\bf x}_{jq}) / 2
    \label{eq:edge_weights}
\end{equation}

Here, ${\bf x}_{ip}$ is the feature value of node $i$ in layer $p$, ${\bf x}_{jq}$ is the feature value of node $j$ in layer $q$, and $w(v_{ip}, v_{jq})$ is the weight of the interlayer link from node $i$ in layer $p$ to node $j$ in layer $q$.

\begin{table}[!ht]
    \centering
    \begin{tabular}{ccccc}
        \hline
        Type & Layers & Nodes & $p$ & $k$ \\
        \hline
        Complete & 2,3& 2-10& - & - \\
        \hline
        Random & 2,3& 4-10& 0.3-0.7& - \\
        \hline
        Small World& 2,3& 4-10& 0.3-0.7& 2,4\\
        \hline
    \end{tabular}
    \caption{The number of nodes and layers used to generate synthetic data. Here, $p$ represents the probability of existence of a link in random networks, and $k$ represents the number of nearest neighbors for each node in Newman-Watts-Strogatz small-world networks.}
    \label{tab:graph_parameters}
\end{table}

\begin{figure}[!ht]
    \centering
    \includegraphics[width=\textwidth,height=0.5\textheight,keepaspectratio]{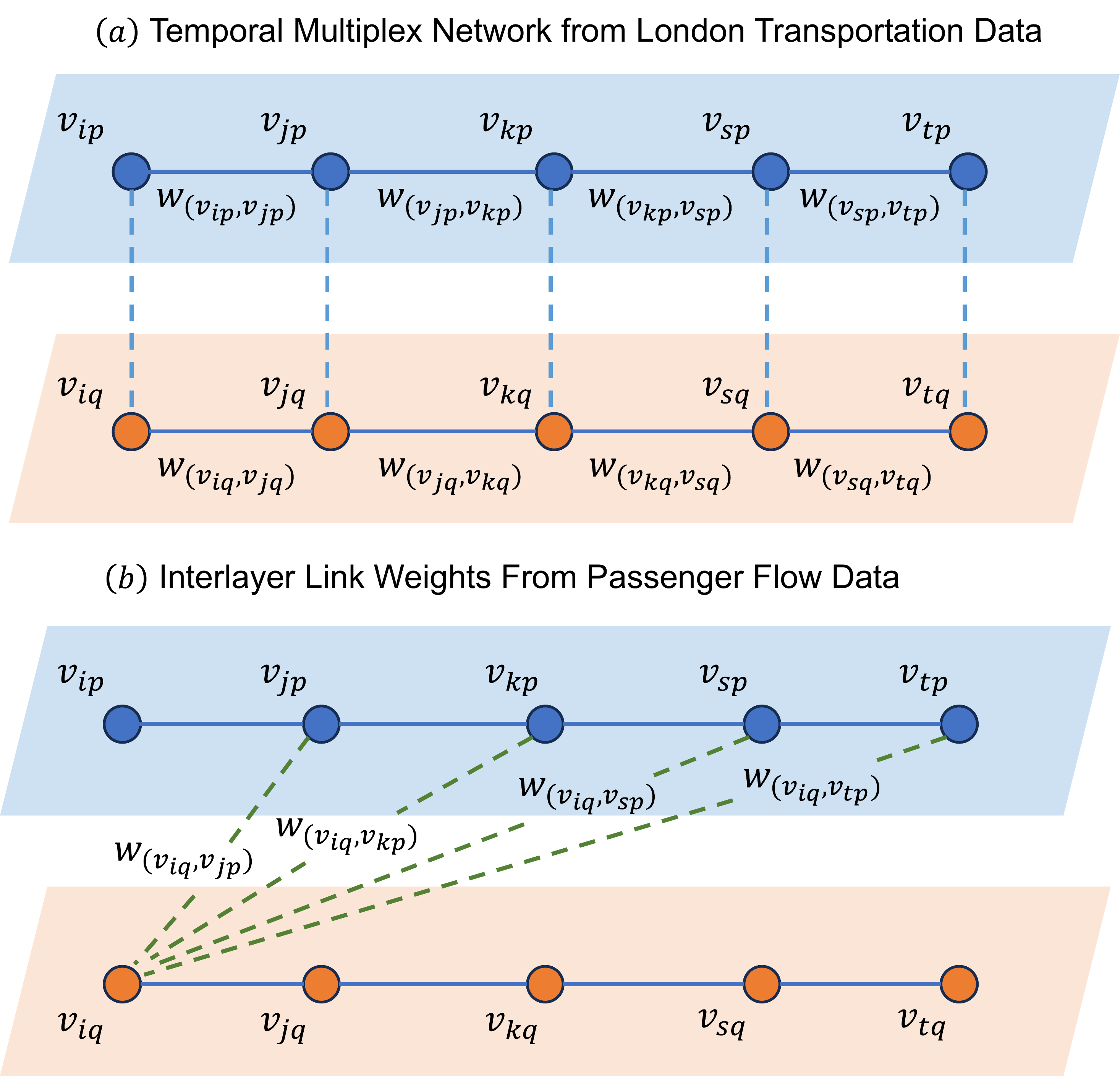}
    \caption{Data preparation pipeline for the London transportation network dataset.  In (a) five connected stations $\{v_i,v_j,v_k,v_s,v_t\}$ from the London Underground were selected to construct temporal multiplex networks, where the two layers $\{p,q\}$ correspond to adjacent 15‑minute intervals. The number of passangers boarding the train at each station were used as node features $\{ \mathbf x_i, \mathbf x_j, \mathbf x_k, \mathbf{x_s}, \mathbf{x_t}\}$ and the number of passengers traveling between stations were used as intralayer link weights.  In (b) the passenger flow between connected stations across time intervals was calculated and used as the interlayer link weights.}
    \label{fig:london_data_preparation}
\end{figure}

\paragraph{London Transportation Dataset:} We also evaluated our method using a real‑world transit dataset provided by Transport for London (TfL) \cite{london_datastore}. We used passenger flow data from 2016-2023 to construct temporal multiplex networks, where each multiplex network contained two layers representing adjacent 15 minute time intervals. For example, a single multiplex network included two layers where layer, \(p\), has data for the 08:00-08:15 time period, and layer, \(q\), has data for 08:15-08:30. The nodes in each layer represent five adjacent stations with non-zero passenger flow in the central line of the London Underground (Bank and Monument, Liverpool Street, Bethnal Green, Mile End, and Stratford).  We then used the number of passengers boarding at each station as node features, the number of passengers traveling between stations as the intralayer link weights, and the passenger flow between stations in different time intervals as the interlayer link weights.  The details of the data preparation are shown in figure \ref{fig:london_data_preparation}, where $v$ are the stations, ${x}$ are the node features, ${w}$ are the link weights between stations, and ${p}$ and ${q}$, represent adjacent layers (15 minute time intervals). 

\subsubsection{Model Training}
We trained the MSGCN model on both synthetic and London transportation data to verify the generalizability of the proposed approach. The details of each set of experiments are described below. For the synthetic data, Five hundred networks were generated for each network type in Table \ref{tab:graph_parameters}, with 400 used for training and the remaining 100 for testing according to an 80/20 split rule. Twenty-five trials were performed for each network type, with network sizes ranging from $2$ to $10$ nodes. Within each trial, the graph convolutional network was trained for $40$ epochs with $32$ hidden layers using a dropout rate of $0.5$. The ADAM optimizer was used with a learning rate of $1 \times 10^{-4}$. In order to evaluate our method in a real world scenario, we predicted passenger flow in 2023 using training data from 2016-2022.  The training data contained 217 networks from 2016-2022 and the testing data used 31 networks from 2023. We performed five independent trials using the same configuration as in the simulated data experiments.


\subsubsection{Validation}
We validated the performance of our Multiplex Spatial Graph Convolution Network (MSGCN) based on four criteria. First, we compared the prediction results to the ground truth in our generated synthetic data using two metrics: R-squared ($R^2$) and Pearson's correlation coefficient ($r$). Next, we compared the performance of our model to other link weight prediction models. Since there are no existing multilayer link weight prediction models, we compared our method to current single-layer link weight prediction models applied to the multilayer case. These extended single-layer models include the Weisfeiler–Lehman algorithm (LWP) \cite{zhang2017weisfeiler}, conventional Graph Convolutional Network (GCN) \cite{zhang2018link}, and Node2Vec \cite{khanam2021noddle}. We then tested the generalization of our proposed method relative to network type and size. Lastly, we employed the London transportation dataset and quantified the absolute error of interlayer link‑weight predictions to demonstrate the method’s ability to generalize to real‑world network data.

\section{Results} \label{Results}

\subsection{Link Weight Prediction Performance using Synthetic Data}
\begin{figure}[!ht]
\centering
\includegraphics[width=.7\textwidth]{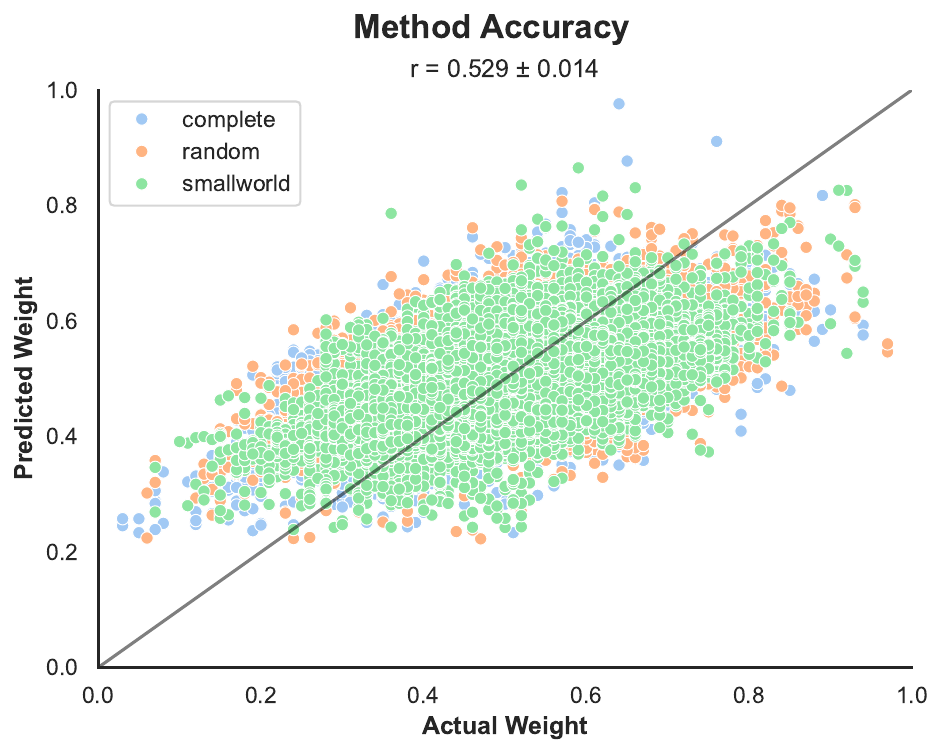}
\caption{Predicted link weights compared with actual values for all synthetic data experiments conducted in this study. The results demonstrate the effectiveness of the MSGCN method in predicting unseen interlayer link weights across various network types, network sizes and number of layers.}
\label{fig:link_weight_predictions}
\end{figure}

Figure \ref{fig:link_weight_predictions} shows the goodness-of-fit plot for all interlayer link weight prediction experiments using the synthetic data. The aggregated results indicate a total Pearson correlation value of $r=0.529 \pm 0.014 $, demonstrating that the model captures a consistent portion of the variance in the true interlayer link weights. The detailed case by case results are provided in the supplementary materials. 

\subsection{Comparison with other Link Weight Prediction Methods}

\begin{figure}[!ht]
\centering 
\includegraphics[width=\textwidth]{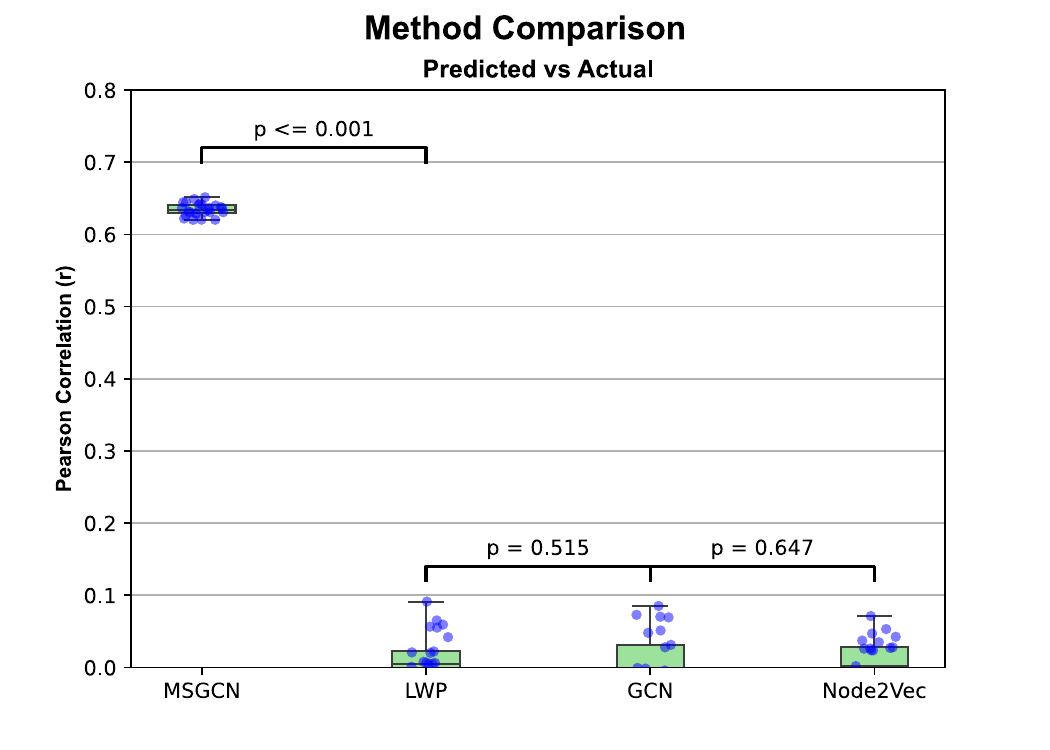}
\caption{A comparison of different interlayer link weight prediction models based on 25 simulated multilayer networks shows that the MSGCN method has significantly higher Pearson correlation values between the predicted values and the ground truth. The p-values correspond to the t-tests that were conducted to compare the means of different groups in the study.}

\label{fig:pcc_comparison}
\end{figure}

We also compared our method to three existing link weight prediction methods (LWP, GCN, and Node2Vec) that were extended to the multilayer case. We repeated the experiments 25 times for each method using the synthetic data, and the Pearson correlation results are shown in Figure \ref{fig:pcc_comparison}. In Addition, the mean squared error comparisons are shown in Table \ref{tab:mse}. The results suggest that the MSGCN method provides significantly better performance compared to the other tested methods. 

\begin{table}[!ht]
  \centering
  \begin{tabular}{ccccc}
    \hline
    Method & Simple & Complete & Random & Small World\\
    \hline
    Node2Vec & $126.75 \pm 3.97$ & $231.89 \pm 3.88$ & $230.98 \pm 6.05$ & $228.69 \pm 6.2$ \\
    \hline
    GCN & $0.117 \pm 0.0317$ & $0.189 \pm .0903$ & $0.0187 \pm .0879$ & $0.184 \pm 0.0844$\\
    \hline
    LWP-WL & $0.285 \pm .0621$ & $0.322 \pm 0.0228$ & $0.323 \pm 0.0326$ &  $0.322 \pm 0.0233$\\
    \hline
    \textbf{MSGCN} & $\mathbf{0.0086 \pm 0.0012}$ & $\mathbf{0.0132 \pm 0.047}$ & $\mathbf{0.0127 \pm .0112}$ & $\mathbf{0.0096 \pm 0.0056}$\\
    \hline
  \end{tabular}
  \caption{The mean squared error between predicted values and the actual interlayer link weights acorss different experiments.}
  \label{tab:mse}
\end{table}

\subsection{Robustness in Terms of Network Size and Type}
\begin{figure}[!ht]
    \centering
    \includegraphics[width=\textwidth] {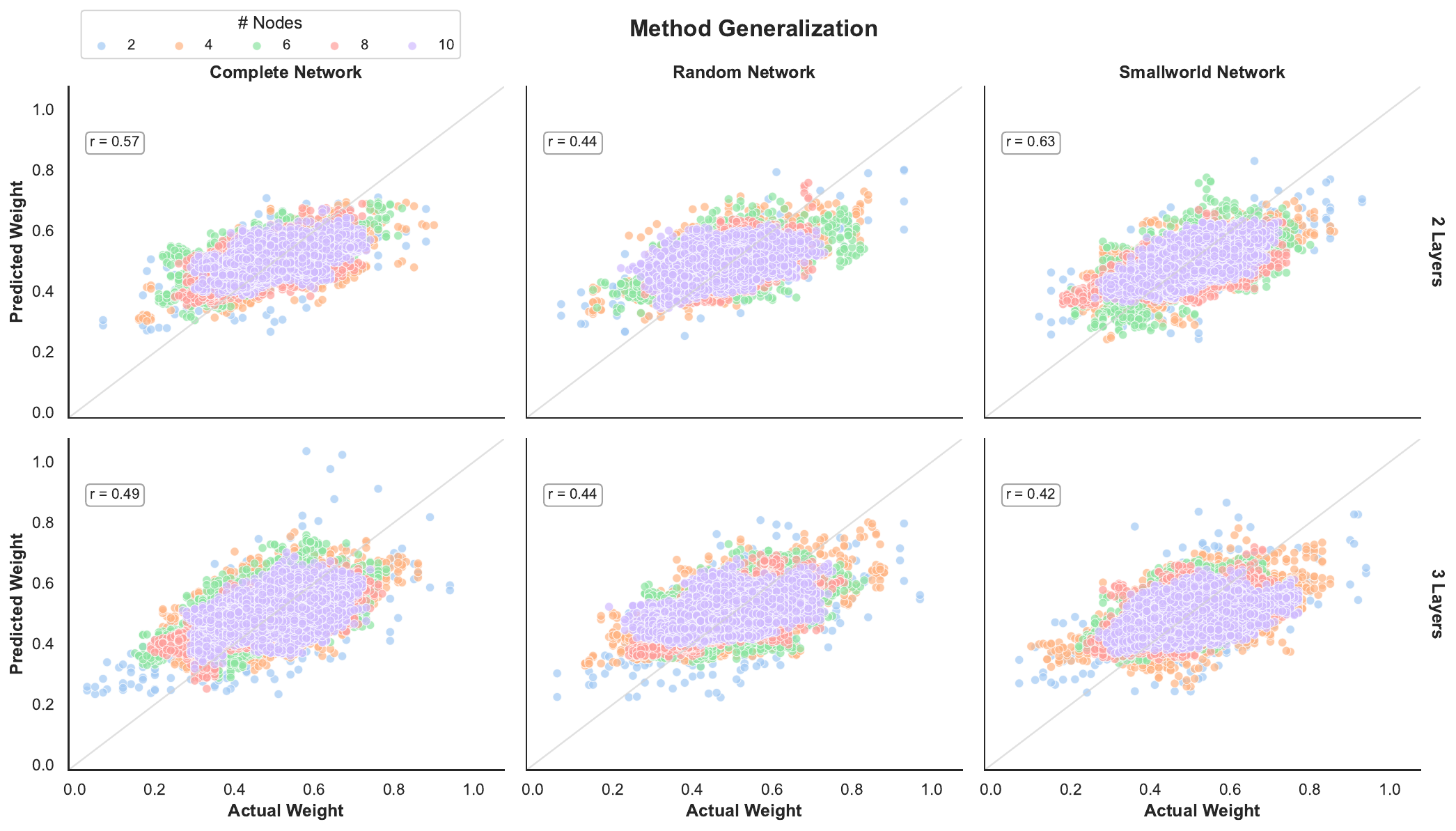}
    \caption{The performance of the MSGCN model was evaluated against different network types and sizes. The results suggest the robustness of MSGCN model across various network configurations and sizes. The r within each plot indicates the average Pearson correlation value.}
    \label{fig:method_performance_all}
\end{figure}

\begin{figure}[!ht]
\centering
\includegraphics[width=\textwidth]{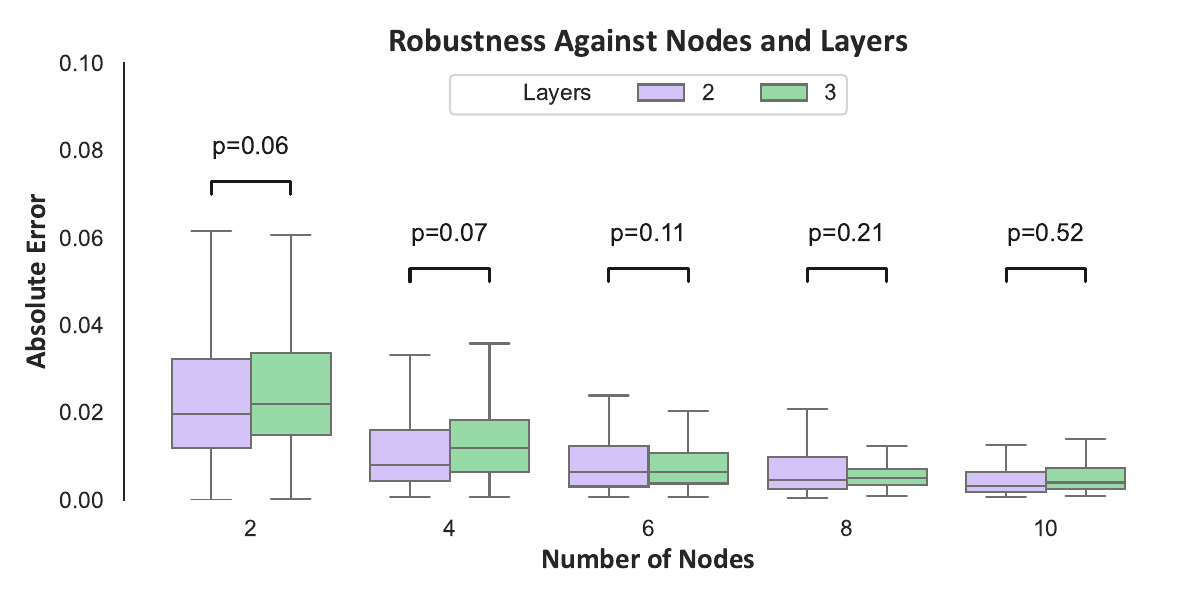}
\caption{Robustness of the MSGCN method against the number of nodes and layers. The y-axis represents the absolute error between predicted interlayer weights and actual values. The p-values correspond to the t-tests between mean of error values.}
\label{fig:robustness_by_nodes_layers}
\end{figure}

\begin{figure}[!ht]
\centering
\includegraphics[width=\textwidth]{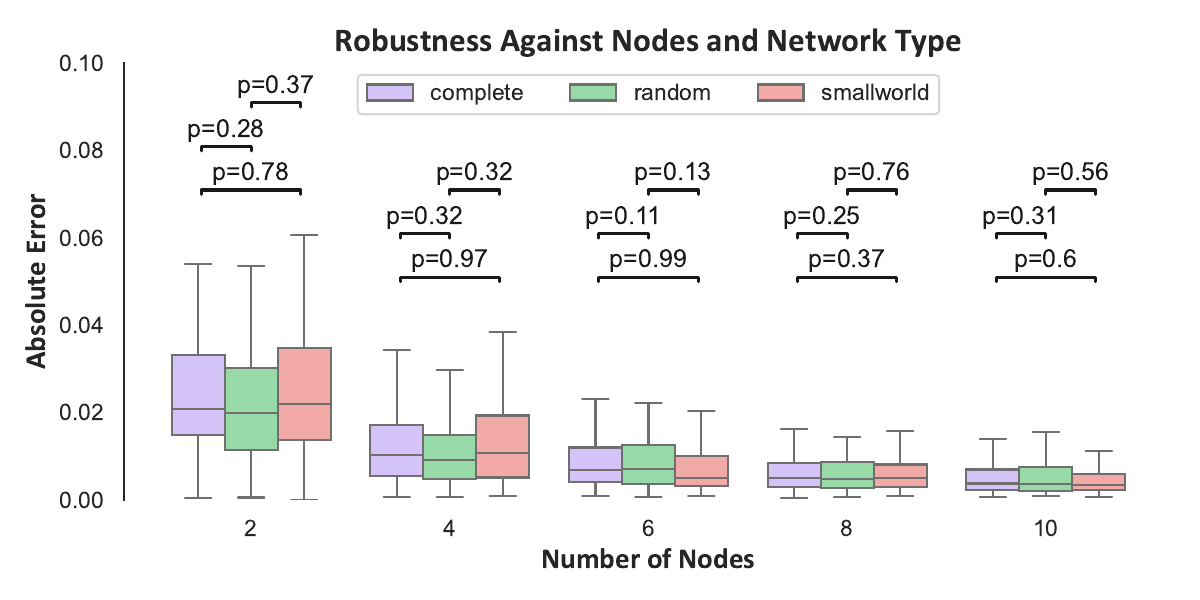}
\caption{Robustness of MSGCN method against number of nodes and network type.The y-axis represents the absolute error between predicted interlayer weights and actual values. The p-values correspond to the t-tests between mean of error values.}
\label{fig:robustness_by_nodes_type}
\end{figure}

\begin{figure}[!ht]
\centering
\includegraphics[width=\textwidth]{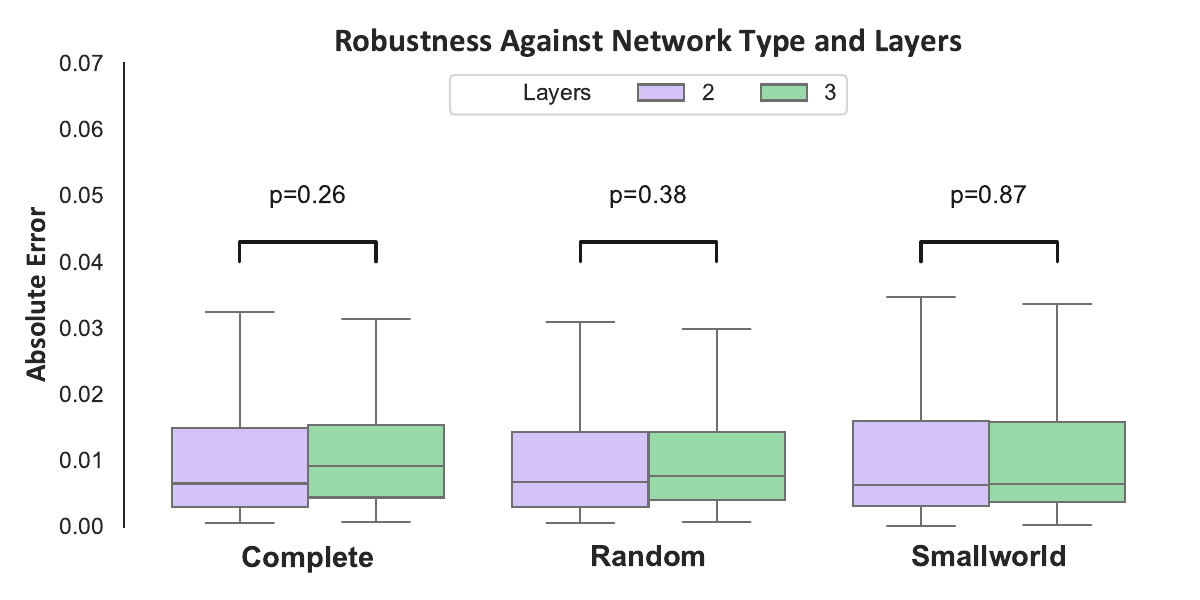}
\caption{Robustness of MSGCN method against number of layers and network type. The y-axis represents the absolute error between predicted interlayer weights and actual values. The p-values correspond to the t-tests between mean of error values.}
\label{fig:robustness_by_layers_type}
\end{figure}

We also tested the robustness of our method across different network types and sizes. Figure \ref{fig:method_performance_all} presents the detailed results of our sensitivity analysis, evaluating the performance of the MSGCN method across various multiplex network types (complete, random, and small-world) with varying numbers of nodes ($2-10$) and layers ($2-3$). The performance of the MSGCN method was consistent across different experiments; nevertheless, the worst performance was in random network configurations, which was expected as these network types lack an inherent structure that makes it more difficult for the prediction models to identify meaningful patterns. We also provided the summary results in terms of the number of nodes versus the number of layers in figure \ref{fig:robustness_by_nodes_layers}, the number of nodes versus network type in figure \ref{fig:robustness_by_nodes_type}, and the number of layers versus network type in figure \ref{fig:robustness_by_layers_type}. All the results suggest that MSGCN maintains robust performance across changes in the number of nodes, the number of layers, and the network type. Nonetheless, prediction error decreases as the number of nodes increases, which is expected since larger networks provide more training samples for the model.

\subsection{Link Weight Prediction Performance on London Transportation Data}
\begin{figure}[!ht]
    \centering
    \includegraphics[width=1.0\linewidth]{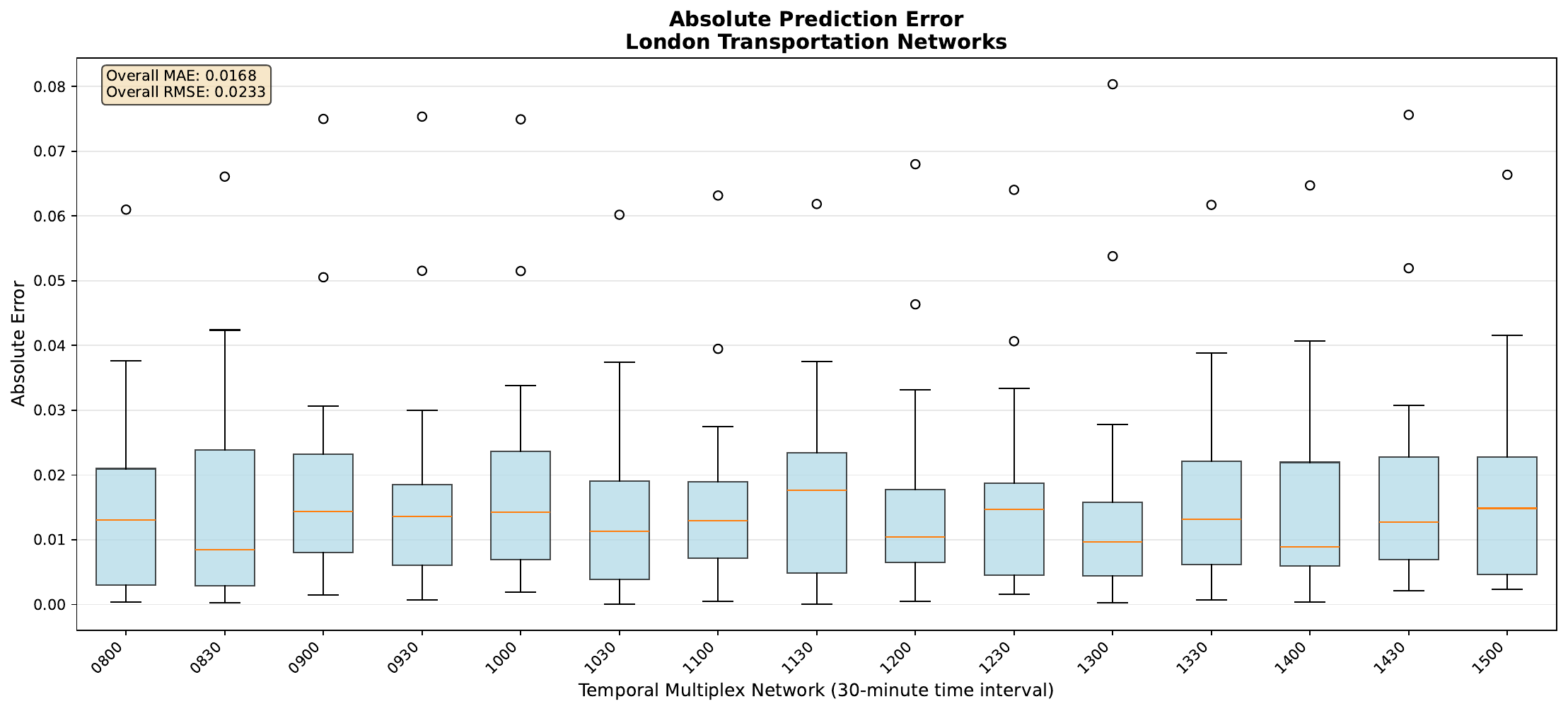}
    \caption{Absolute error of interlayer link‑weight predictions across time intervals in the London dataset. Each boxplot represents the distribution of absolute errors for predictions made over 30‑minute intervals. The results highlight the MSGCN method’s low prediction error and robustness in real‑world settings.}
    \label{fig:london_accuracy}
\end{figure}

In order to showcase the real‑world applicability of the MSGCN method, we tested it on the London Transportation network dataset \cite{london_datastore}. We used the Transport for London (TfL) Project NUMBAT dataset on usage and travel patterns to simulate a capacity planning activity.  We trained the model using traffic data from Saturdays between 2016 and 2022 for five connected stations in the London Underground. We then predicted the traffic flow across different stations on Saturdays in 2023. Figure \ref{fig:london_accuracy} shows the predicted and actual interlayer link weights using MSGCN method.  

\section{Discussion} \label{Discussion}

This study addresses an important gap in multilayer network analysis, specifically the link weight prediction problem, where the goal is to estimate the strength of interlayer connections in the network. To the best of our knowledge, this is one of the first methods proposed for interlayer link weight prediction that directly embeds node features from different layers of a network. Furthermore, our proposed method takes into account the spatial location of nodes during graph embedding, making it applicable to many real-world complex systems where the nodes have spatial properties. Lastly, we have shown that our proposed method generalizes to different network sizes and structures while mitigating the oversmoothing problem often seen in graph neural networks.

\subsection{Scaling to very Large Graphs}
An important practical aspect of any graph-based method is its ability to scale to very large networks. The time complexity of the MSGCN algorithm is based on its core operations, including the node feature projection and spatial graph convolution. Assuming the number of nodes is $n$ and the number of layers is $m$, the node projection has a time complexity of $O(nm)$. In practice, the number of layers is much smaller than the number of nodes, so the complexity of the projection step becomes linear with respect to the number of nodes, $O(n)$. The complexity of the graph convolution step is dominated by the number of edges $|E|$. In the worst case of fully connected (complete) graphs where $|E| \sim \frac{n^2}{2}$, the complexity approaches $O(n^2)$. Hence, in the worst‑case scenario, the overall time complexity of the MSGCN algorithm becomes quadratic with respect to the number of nodes, $O(n^2)$. This complexity indicates that, while the algorithm is efficient for networks with a moderate number of nodes, its performance could be affected in large multilayer networks. To enhance the scalability of the MSGCN method, our future work will explore more efficient optimization techniques based on sparse matrix operations, parallel processing, and approximation techniques for the convolution operations.

\subsection{Generalizing Beyond Multiplex Networks}
Another practical consideration is the ability of our method to be applied to a diverse range of real-world problems. In this work, we demonstrated that the MSGCN method is generalizable to multiple network types with different topologies. In addition, we also demonstrated that the MSGCN method can be applied to temporal networks by treating them as multiplex networks, where each layer corresponds to a different time interval. However, we mainly focused on multiplex networks, which are a specific type of multilayer network that share the same set of nodes across all layers. Extending the proposed method to heterogeneous multilayer networks, in which layers may vary in structure or contain different node sets, could enable broader practical applications and will be explored in future work. Furthermore, the proposed method is based on supervised learning and relies on known interlayer link weights for training. In practice, we may not have access to such information, so an unsupervised extension of the proposed method would be necessary to broaden its application.

\section{Conclusion} \label{Conclusion}

In this paper, we introduced a new method termed Multiplex Spatial Graph Convolution Network (MSGCN) to predict the interlayer link weights in multiplex networks. The method works by projecting features and embedding them spatially across multiple layers of a network. Our results show that MSGCN can accurately predict the interlayer link weights and outperforms other link weight prediction models in synthetic data with known link weights. Our results also show that the proposed method is accurate, robust, and generalizable across a wide variety of multiplex networks. The MSGCN method can be used in several real-world scenarios ranging from biological to transportation networks. We demonstrated a real‑world application of the proposed method using the London transportation data, where we constructed a temporal multiplex network to predict the passenger flow between stations. 
Taken together, these findings highlight the effectiveness of MSGCN method for analyzing complex multilayer systems. In future work, we plan to extend the method to more heterogeneous multilayer networks and utilize it to provide new insights into their complex behavior in real‑world scenarios.


 
\section{Code and Data availability}
All the code and data used in this study are available at: \\
\href{https://github.com/3sigmalab/MSGCN}{https://github.com/3sigmalab/MSGCN}

\bibliographystyle{comnet}
\bibliography{references}

@misc{london_datastore,
    title ="London TfL Open Data" ,
    author="Transport for London",
    url="https://crowding.data.tfl.gov.uk/#!NUMBAT%2"
}

@misc{Liu_2017,
      title={Principled Multilayer Network Embedding}, 
      author={Weiyi Liu and Pin-Yu Chen and Sailung Yeung and Toyotaro Suzumura and Lingli Chen},
      year={2017},
      eprint={1709.03551},
      archivePrefix={arXiv},
      primaryClass={cs.SI}
}

@ARTICLE{Kivela_2014,
  title     = "Multilayer networks",
  author    = "Kivela, M and Arenas, A and Barthelemy, M and Gleeson, J P and
               Moreno, Y and Porter, M A",
  journal   = "J. Complex Netw.",
  publisher = "Oxford University Press (OUP)",
  volume    =  2,
  number    =  3,
  pages     = "203--271",
  month     =  sep,
  year      =  2014,
  language  = "en"
}

@misc{danel_2020,
      title={Spatial Graph Convolutional Networks}, 
      author={Tomasz Danel and Przemysław Spurek and Jacek Tabor and Marek Śmieja and Łukasz Struski and Agnieszka Słowik and Łukasz Maziarka},
      year={2020},
      eprint={1909.05310},
      archivePrefix={arXiv},
      primaryClass={cs.LG}
}

@misc{kipf_2017,
      title={Semi-Supervised Classification with Graph Convolutional Networks}, 
      author={Thomas N. Kipf and Max Welling},
      year={2017},
      eprint={1609.02907},
      archivePrefix={arXiv},
      primaryClass={cs.LG}
}

@misc{kipf_2016,
      title={Variational Graph Auto-Encoders}, 
      author={Thomas N. Kipf and Max Welling},
      year={2016},
      eprint={1611.07308},
      archivePrefix={arXiv},
      primaryClass={stat.ML}
}

@misc{rusch_2023,
      title={A Survey on Oversmoothing in Graph Neural Networks}, 
      author={T. Konstantin Rusch and Michael M. Bronstein and Siddhartha Mishra},
      year={2023},
      eprint={2303.10993},
      archivePrefix={arXiv},
      primaryClass={cs.LG},
      url={https://arxiv.org/abs/2303.10993}, 
}

@misc{hamilton_2018,
      title={Representation Learning on Graphs: Methods and Applications}, 
      author={William L. Hamilton and Rex Ying and Jure Leskovec},
      year={2018},
      eprint={1709.05584},
      archivePrefix={arXiv},
      primaryClass={cs.SI},
      url={https://arxiv.org/abs/1709.05584}, 
}

@inproceedings{Wu_2019,
author = {Wu, Jun and He, Jingrui and Xu, Jiejun},
title = {DEMO-Net: Degree-specific Graph Neural Networks for Node and Graph Classification},
year = {2019},
isbn = {9781450362016},
publisher = {Association for Computing Machinery},
address = {New York, NY, USA},
url = {https://doi.org/10.1145/3292500.3330950},
doi = {10.1145/3292500.3330950},
booktitle = {Proceedings of the 25th ACM SIGKDD International Conference on Knowledge Discovery \& Data Mining},
pages = {406–415},
numpages = {10},
location = {Anchorage, AK, USA},
series = {KDD '19}
}

@inproceedings{ren2024link,
  title={Link Prediction in Multilayer Networks via Cross-Network Embedding},
  author={Ren, Guojing and Ding, Xiao and Xu, Xiao-Ke and Zhang, Hai-Feng},
  booktitle={Proceedings of the AAAI Conference on Artificial Intelligence},
  volume={38},
  number={8},
  pages={8939--8947},
  year={2024}
}

@article{yang2024link,
  title={Link prediction for multi-layer and heterogeneous cyber-physical networks},
  author={Yang, Guoli and Liu, Yi},
  journal={International Journal of Machine Learning and Cybernetics},
  pages={1--17},
  year={2024},
  publisher={Springer}
}

@article{aleta2017multilayer,
  title={A multilayer perspective for the analysis of urban transportation systems},
  author={Aleta, Alberto and Meloni, Sandro and Moreno, Yamir},
  journal={Scientific reports},
  volume={7},
  number={1},
  pages={44359},
  year={2017},
  publisher={Nature Publishing Group UK London}
}

@article{wu2020traffic,
  title={Traffic dynamics on multilayer networks},
  author={Wu, Jiexin and Pu, Cunlai and Li, Lunbo and Cao, Guo},
  journal={Digital Communications and Networks},
  volume={6},
  number={1},
  pages={58--63},
  year={2020},
  publisher={Elsevier}
}

@article{vaiana2020multilayer,
  title={Multilayer brain networks},
  author={Vaiana, Michael and Muldoon, Sarah Feldt},
  journal={Journal of Nonlinear Science},
  volume={30},
  number={5},
  pages={2147--2169},
  year={2020},
  publisher={Springer}
}

@article{liu2024motifs,
  title={Motifs-based link prediction for heterogeneous multilayer networks},
  author={Liu, Yafang and Zhou, Jianlin and Zeng, An and Fan, Ying and Di, Zengru},
  journal={Chaos: An Interdisciplinary Journal of Nonlinear Science},
  volume={34},
  number={9},
  year={2024},
  publisher={AIP Publishing}
}

@article{la2025heuristic,
  title={Heuristic-Informed Mixture of Experts for Link Prediction in Multilayer Networks},
  author={La Cava, Lucio and Mandaglio, Domenico and Zangari, Lorenzo and Tagarelli, Andrea},
  journal={arXiv preprint arXiv:2501.17557},
  year={2025}
}

@inproceedings{zangari2024link,
  title={Link Prediction on Multilayer Networks through Learning of Within-Layer and Across-Layer Node-Pair Structural Features and Node Embedding Similarity},
  author={Zangari, Lorenzo and Mandaglio, Domenico and Tagarelli, Andrea},
  booktitle={Proceedings of the ACM on Web Conference 2024},
  pages={924--935},
  year={2024}
}

@article{liu2024tlfsl,
  title={TLFSL: link prediction in multilayer social networks using trustworthy L{\'e}vy-flight semi-local random walk},
  author={Liu, Mingchun and Jannesari, Vahid},
  journal={Journal of Complex Networks},
  volume={12},
  number={4},
  pages={cnae026},
  year={2024},
  publisher={Oxford University Press}
}

@article{luo2024link,
  title={Link prediction in multilayer networks using weighted reliable local random walk algorithm},
  author={Luo, Zhiping and Yin, Jian and Lu, Guangquan and Rahimi, Mohammad Reza},
  journal={Expert Systems with Applications},
  volume={247},
  pages={123304},
  year={2024},
  publisher={Elsevier}
}

@article{jafari2021information,
  title={An information theoretic approach to link prediction in multiplex networks},
  author={Jafari, Seyed Hossein and Abdolhosseini-Qomi, Amir Mahdi and Asadpour, Masoud and Rahgozar, Maseud and Yazdani, Naser},
  journal={Scientific Reports},
  volume={11},
  number={1},
  pages={13242},
  year={2021},
  publisher={Nature Publishing Group UK London}
}

@article{aleta2020link,
  title={Link prediction in multiplex networks via triadic closure},
  author={Aleta, Alberto and Tuninetti, Marta and Paolotti, Daniela and Moreno, Yamir and Starnini, Michele},
  journal={Physical Review Research},
  volume={2},
  number={4},
  pages={042029},
  year={2020},
  publisher={APS}
}

@article{baptista2022universal,
  title={Universal multilayer network exploration by random walk with restart},
  author={Baptista, Anthony and Gonzalez, Aitor and Baudot, Ana{\"\i}s},
  journal={Communications Physics},
  volume={5},
  number={1},
  pages={170},
  year={2022},
  publisher={Nature Publishing Group UK London}
}

@article{wangmultiple2025,
author = {Wang, Huan and Teng, Yu and Qin, Lingsong and Guo, Xuan and Hu, Po}, title = {A Multiple Attention Layer-shareable Method for Link Prediction in Multilayer Networks}, year = {2025}, issue_date = {February 2025}, publisher = {Association for Computing Machinery}, address = {New York, NY, USA}, volume = {19}, number = {2}, issn = {1556-4681}, url = {https://doi.org/10.1145/3709142}, doi = {10.1145/3709142}, journal = {ACM Trans. Knowl. Discov. Data}, month = feb, articleno = {36}, numpages = {22} }

@article{finn2021multilayer,
  title={Multilayer network analyses as a toolkit for measuring social structure},
  author={Finn, Kelly R},
  journal={Current Zoology},
  volume={67},
  number={1},
  pages={81--99},
  year={2021},
  publisher={Oxford University Press}
}

@article{gao2023novel,
  title={A novel link prediction model in multilayer online social networks using the development of Katz similarity metric},
  author={Gao, Zhie and Rezaeipanah, Amin},
  journal={Neural Processing Letters},
  volume={55},
  number={4},
  pages={4989--5011},
  year={2023},
  publisher={Springer}
}

@inproceedings{zhang2017weisfeiler,
  title={Weisfeiler-lehman neural machine for link prediction},
  author={Zhang, Muhan and Chen, Yixin},
  booktitle={Proceedings of the 23rd ACM SIGKDD international conference on knowledge discovery and data mining},
  pages={575--583},
  year={2017}
}

@article{zhang2018link,
  title={Link prediction based on graph neural networks},
  author={Zhang, Muhan and Chen, Yixin},
  journal={Advances in neural information processing systems},
  volume={31},
  year={2018}
}

@inproceedings{khanam2021noddle,
  title={NODDLE: Node2vec based deep learning model for link prediction},
  author={Khanam, Kazi Zainab and Singhal, Aditya and Mago, Vijay},
  booktitle={National Conference on Big Data Technology and Applications},
  pages={196--212},
  year={2021},
  organization={Springer}
}

@article{boccaletti2014structure,
  title={The structure and dynamics of multilayer networks},
  author={Boccaletti, Stefano and Bianconi, Ginestra and Criado, Regino and Del Genio, Charo I and G{\'o}mez-Gardenes, Jes{\'u}s and Romance, Miguel and Sendina-Nadal, Irene and Wang, Zhen and Zanin, Massimiliano},
  journal={Physics reports},
  volume={544},
  number={1},
  pages={1--122},
  year={2014},
  publisher={Elsevier}
}

@article{alessandretti2023multimodal,
  title={Multimodal urban mobility and multilayer transport networks},
  author={Alessandretti, Laura and Natera Orozco, Luis Guillermo and Saberi, Meead and Szell, Michael and Battiston, Federico},
  journal={Environment and Planning B: Urban Analytics and City Science},
  volume={50},
  number={8},
  pages={2038--2070},
  year={2023},
  publisher={SAGE Publications Sage UK: London, England}
}

@article{lehman2011multilayer,
  title={Multilayer networks: An architecture framework},
  author={Lehman, Tom and Yang, Xi and Ghani, Nasir and Gu, Feng and Guok, Chin and Monga, Inder and Tierney, Brian},
  journal={IEEE Communications Magazine},
  volume={49},
  number={5},
  pages={122--130},
  year={2011},
  publisher={IEEE}
}

@article{dragic2021multilayer,
  title={Multilayer social networks reveal the social complexity of a cooperatively breeding bird},
  author={Dragi{\'c}, Nikola and Keynan, Oded and Ilany, Amiyaal},
  journal={Iscience},
  volume={24},
  number={11},
  year={2021},
  publisher={Elsevier}
}

@article{liu2020robustness,
  title={Robustness and lethality in multilayer biological molecular networks},
  author={Liu, Xueming and Maiorino, Enrico and Halu, Arda and Glass, Kimberly and Prasad, Rashmi B and Loscalzo, Joseph and Gao, Jianxi and Sharma, Amitabh},
  journal={Nature communications},
  volume={11},
  number={1},
  pages={6043},
  year={2020},
  publisher={Nature Publishing Group UK London}
}

@article{de2023more,
  title={More is different in real-world multilayer networks},
  author={De Domenico, Manlio},
  journal={Nature Physics},
  volume={19},
  number={9},
  pages={1247--1262},
  year={2023},
  publisher={Nature Publishing Group UK London}
}

@article{wu2023demystifying,
  title={Demystifying oversmoothing in attention-based graph neural networks},
  author={Wu, Xinyi and Ajorlou, Amir and Wu, Zihui and Jadbabaie, Ali},
  journal={Advances in Neural Information Processing Systems},
  volume={36},
  pages={35084--35106},
  year={2023}
}

@article{aleta2019multilayer,
  title={Multilayer networks in a nutshell},
  author={Aleta, Alberto and Moreno, Yamir},
  journal={Annual Review of Condensed Matter Physics},
  volume={10},
  number={1},
  pages={45--62},
  year={2019},
  publisher={Annual Reviews}
}

@article{lu2011link,
  title={Link prediction in complex networks: A survey},
  author={L{\"u}, Linyuan and Zhou, Tao},
  journal={Physica A: Statistical Mechanics and its Applications},
  volume={390},
  number={6},
  pages={1150--1170},
  year={2011},
  publisher={Elsevier},
  doi={10.1016/j.physa.2010.11.027}
}

@article{ding2024wepred,
  title={WePred: Edge Weight-Guided Contrastive Learning for Bipartite Link Prediction},
  author={Ding, Linlin and Han, Yiming and Li, Mo and Gu, Yinghao and Liu, Tingting and Yu, Shidong},
  journal={Electronics},
  volume={14},
  number={1},
  pages={20},
  year={2024},
  publisher={MDPI}
}

@article{iddianozie2020improved,
  title={Improved graph neural networks for spatial networks using structure-aware sampling},
  author={Iddianozie, Chidubem and McArdle, Gavin},
  journal={ISPRS International Journal of Geo-Information},
  volume={9},
  number={11},
  pages={674},
  year={2020},
  publisher={MDPI}
}

@article{zhu2022spatial,
  title={Spatial regression graph convolutional neural networks: A deep learning paradigm for spatial multivariate distributions},
  author={Zhu, Di and Liu, Yu and Yao, Xin and Fischer, Manfred M},
  journal={GeoInformatica},
  volume={26},
  number={4},
  pages={645--676},
  year={2022},
  publisher={Springer}
}

\end{document}